%% file: main.tex
\documentclass[10pt,twocolumn,letterpaper]{article}

\usepackage{iccv}              

\input{preamble}

\definecolor{iccvblue}{rgb}{0.21,0.49,0.74}
\usepackage[pagebackref,breaklinks,colorlinks,allcolors=iccvblue]{hyperref}

\def\paperID{12365} 
\def\confName{ICCV}
\def\confYear{2025}

\title{An Information-Theoretic Regularizer for Lossy Neural Image Compression}

\author{
Yingwen Zhang\textsuperscript{1}, 
Meng Wang\textsuperscript{2}, 
Xihua Sheng\textsuperscript{1}, 
Peilin Chen\textsuperscript{1}, \\
Junru Li\textsuperscript{3},
Li Zhang\textsuperscript{4}, 
Shiqi Wang\textsuperscript{1}\thanks{Corresponding author (email:\textit{shiqwang@cityu.edu.hk})}\\
\textsuperscript{1}Department of Computer Science, City University of Hong Kong, Hong Kong, China \\
\textsuperscript{2}School of Data Science, Lingnan University, Hong Kong, China \\
\textsuperscript{3}Douyin Group (HK) Limited, Hong Kong, China \\
\textsuperscript{4}Bytedance Inc., San Diego, USA \\
}

\begin{document}
\maketitle
\input{sec/0_abstract}    
\input{sec/1_intro}
\input{sec/2_related}

\input{sec/3_method}
\input{sec/4_experiment}
\input{sec/5_conclusion}
{
    \small
    \bibliographystyle{ieeetr}
    \bibliography{main}
}

\input{sec/X_suppl}

\end{document}

%% file: preamble.tex
\usepackage{cases}
\usepackage{bm}
\usepackage{amsthm}
\newtheorem{Theorem}{Theorem}
\newtheorem{Lemma}{Lemma}
\usepackage{algorithm}
\usepackage{algpseudocode}
\usepackage{multirow}


%% file: sec/0_abstract.tex
\begin{abstract}
Lossy image compression networks aim to minimize the latent entropy of images while adhering to specific distortion constraints. However, optimizing the neural network can be challenging due to its nature of learning quantized latent representations. In this paper, our key finding is that minimizing the latent entropy is, to some extent, equivalent to maximizing the conditional source entropy, an insight that is deeply rooted in information-theoretic equalities. Building on this insight, we propose a novel structural regularization method for the neural image compression task by incorporating the negative conditional source entropy into the training objective, such that both the optimization efficacy and the model's generalization ability can be promoted. The proposed information-theoretic regularizer is interpretable, plug-and-play, and imposes no inference overheads. Extensive experiments demonstrate its superiority in regularizing the models and further squeezing bits from the latent representation across various compression structures and unseen domains.
\end{abstract}

%% file: sec/1_intro.tex
\section{Introduction}
Lossy image compression, or more broadly, lossy data compression, is grounded in a fundamental theorem in information theory: the rate-distortion theorem~\cite{shannon1959coding}. Essentially, this theorem establishes a theoretical lower bound for lossy compression and provides valuable guidance for designing practical compression algorithms~\cite{sayood2017introduction,yang2023introduction}.

\begin{figure}[t]
\centering
\includegraphics[width=0.45\textwidth]{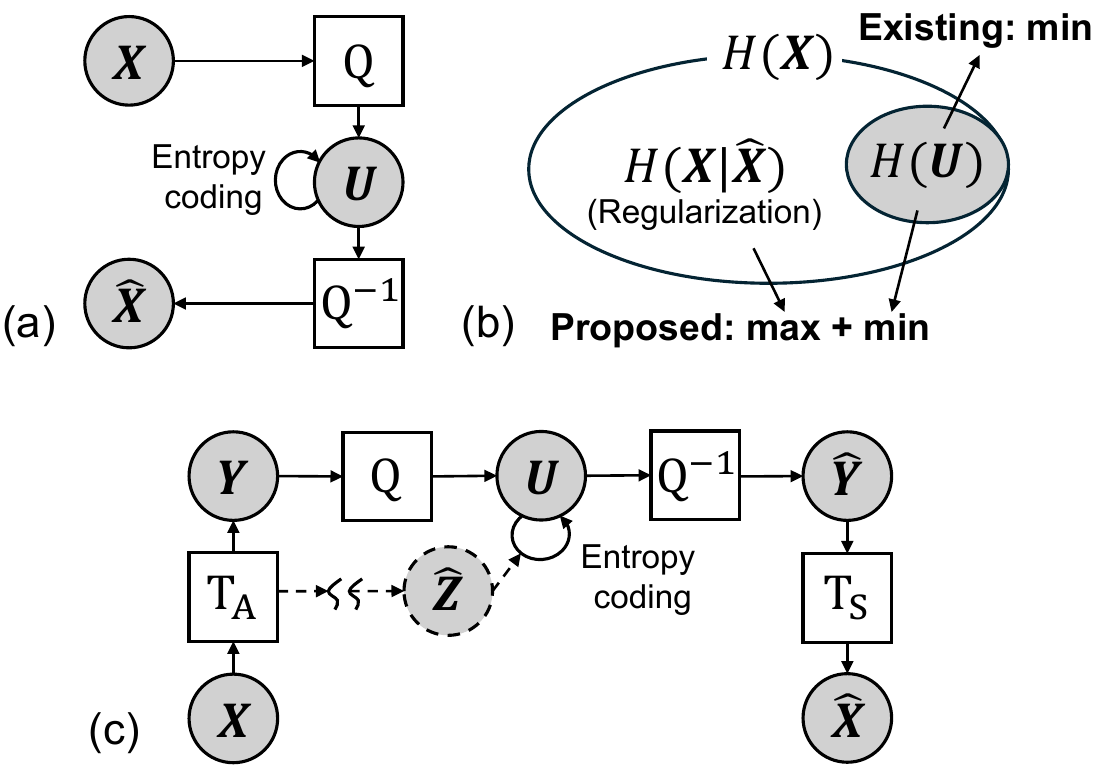}
\caption{(a) Direct coding model; (b) Information diagram for the direct coding, wherein $H(\bm{X})=H(\bm{X}|\bm{\hat{X}})+H(\bm{U})$ and $H(\bm{X})$ is fixed for any known source; (c) Transform coding model (with possible side information $\bm{\hat{Z}}$~\citep{balle2018variational}).}
\label{fig.compression_model}
\end{figure}

A basic lossy image compression model is illustrated in Fig.\ref{fig.compression_model}(a). Let $\bm{X}$, $\bm{\hat{X}}$, and $\bm{U}$ denote the source, the reconstruction, and the discrete index (or the latent), respectively. The encoder is a deterministic mapping $\mathrm{Q}$: $\mathcal{X}^N\rightarrow\mathcal{U}$, where $\mathcal{X}^N$ denotes a $N$-dimensional image space and $\mathcal{U}=\{1,2,...,M\}$ denotes the discrete index space. The decoder is a bijection mapping $\mathrm{Q}^{-1}$: $\mathcal{U}\rightarrow\mathcal{\hat{X}}^N$, where the reconstruction space $\mathcal{\hat{X}}^N=\{\bm{c}_1,...,\bm{c}_M\}$ forms a $N$-dimensional codebook. For index $\bm{U}$, entropy coding is applied to losslessly convert the data into the bitstream, which can then be decoded back to $\bm{U}$ on the decoder side. 
We may refer to such a model as the direct coding model for its structural simplicity. Subsequently, following the rate-distortion theorem~\cite{shannon1959coding}, given a bounded distortion measure $d(\cdot)$ and a distortion constraint $D$, the rate of transmitting $\bm{U}$ is lower bounded by the informational rate-distortion function: 
\begin{align}\label{eq:RD_Function}
R(D)=\mathop{\text{min}}\limits_{p_{\bm{\hat{X}}|\bm{X}}:\mathbb{E}_{ p_{\bm{X}}^{}p_{\bm{\hat{X}}|\bm{X}}}{[d(\bm{X},\bm{\hat{X}})]}<D}\ {\frac{1}{N}I(\bm{X};\bm{\hat{X}})},
\end{align}
i.e., the minimum mutual information obtained by traversing all the possible encoder-decoder (or quantizer) designs $p_{\bm{\hat{X}}|\bm{X}}$. Theoretically, this bound can be achieved asymptotically using the vector quantization as $N\rightarrow\infty$~\cite{berger2003rate}. However, since the encoding complexity grows exponentially with the image dimension $N$, a suboptimal yet more practical solution is the transform coding~\cite{goyal2001theoretical}. As depicted in Fig.\ref{fig.compression_model}(c), for the transform coding model, a deterministic analysis transform $\mathrm{T}_\mathrm{A}$: $\mathcal{X}^N\rightarrow\mathcal{Y}^M$ and a deterministic synthesis transform $\mathrm{T}_\mathrm{S}$: $\mathcal{\hat{Y}}^M\rightarrow\mathcal{\hat{X}}^N$ are incorporated into the compression chain. The analysis transform is designed to decorrelate the source $\bm{X}$, such that a much simpler quantization scheme, e.g., uniform scalar quantization: $\bm{u}_i = \mathrm{round}(\bm{y}_i)$, can be independently applied to each dimension of the transformed coefficients $\bm{Y}$. Building on this framework and leveraging the superior nonlinear capabilities of neural networks, numerous neural image compression networks~\cite{balle2016end,balle2018variational,minnen2018joint,minnen2020channel,cheng2020learned, he2021checkerboard,jiang2023mlic,qin2024mambavc} are proposed, continually pushing the frontier of lossy image compression.

Typically, by parameterizing the $\mathrm{T}_\mathrm{A}$ and $\mathrm{T}_\mathrm{S}$ in Fig.\ref{fig.compression_model}(c) as neural networks, the rate-distortion training objective is formulated as~\cite{balle2016end}:
\begin{gather}\label{eq:rate-dist-loss}
    \mathop{\text{min}}\ \underbrace{{\mathbb{E}_{\bm{X}}[-\text{log}\ q_{\phi}(\bm{U})]}}_{R\approx H(\bm{U})}+\lambda \underbrace{\mathbb{E}_{\bm{X}}||\bm{X}-\bm{\hat{X}}||_2^2,}_{D}
\end{gather}
where $R$ is the estimation of latent entropy $H(\bm{U})$; $q_{\phi}$ is the neural entropy model (or estimated distribution) for $\bm{U}$, whose modeling accuracy can be significantly enhanced by transmitting a small amount of side information $\bm{\hat{Z}}$ (as depicted in Fig.\ref{fig.compression_model}(c)) as part of the parameters $\phi$~\cite{balle2018variational}; the distortion $D$ is the $l_2$ distance between $\bm{X}$ and $\bm{\hat{X}}$; $\lambda$ is the Lagrangian multiplier, controlling the trade-off between $R$ and $D$. Herein, due to the existence of quantization $\mathrm{Q}(\cdot)$, $\bm{U}$ is discrete in essence. Optimizing the network is then challenging as the gradient is almost zero everywhere for $\mathrm{Q}(\cdot)$. To effectively train the networks, a surrogate is introduced during training, e.g., additive uniform noise (AUN)~\cite{balle2016end}:
\begin{gather}\label{eq:AUN}
    \bm{U}=\mathrm{Q}(\bm{Y})\approx\bm{Y}+[-0.5,0.5]^M,
\end{gather}
such that the backward gradient of $\mathrm{Q}(\cdot)$ is one. However, since such approximation inevitably results in the train-test mismatch, and thus high gradient estimation error~\cite{zhang2023uniform}, both the efficiency and efficacy of the optimization are hindered. To address this, many research efforts are devoted to designing advanced surrogates and optimization methods~\cite{agustsson2017soft,yang2020improving,agustsson2020universally,guo2021soft,zhang2023uniform}.

In this paper, we present a completely distinct view on facilitating the optimization of the compression network: regularization. Conceivably, imposing structural constraints on the network optimization can effectively regularize the gradient estimation error, promote convergence, and reduce overfitting. In the context of generative variational autoencoders~\cite{kingma2019introduction}, structural regularization techniques~\cite{zhao2018adversarially,xu2020learning,wu2020vector,le2018supervised,ma2018constrained,sinha2021consistency} have been effectively applied to improve authentic data generation. However, since neural compression networks fundamentally differ from generative autoencoders in terms of the minimum latent entropy constraint, to the best of our knowledge, no structural regularization methods tailored to this constraint have been explored. In light of this, in this paper, we present the first study of information-theoretic regularization for the neural image compression task. In particular, through an in-depth information-theoretic analysis of compression models, we reveal that minimizing the latent entropy $H(\bm{U})$ is, to some extent, equally important as maximizing the conditional source entropy $H(\bm{X}|\bm{\hat{X}})$. Subsequently, by enforcing the network to simultaneously minimize $-H(\bm{X}|\bm{\hat{X}})$, both the in-domain compact representation and out-of-domain generalization ability are experimentally improved.


%% file: sec/2_related.tex
\section{Related works}
\paragraph{Neural image compression networks.} Neural image compression networks are primarily composed of transform, quantization, and latent entropy models. The transform module maps the input source into a latent representation so that the dependencies of the source domain can be more easily handled. Unlike the linear transform~\cite{ochoa2019discrete} used in traditional hybrid coding, the neural transform is inherently nonlinear, enabling more effective decorrelation of the source data. In the literature, various neural transform architectures are proposed. Representative architectures include the recurrent neural network (RNN)~\cite{toderici2015variable}, convolutional neural network (CNN)~\cite{balle2016end}, non-local attention~\cite{liu2019non}, transformer~\cite{lu2021transformer}, mixed transformer-CNN~\cite{liu2023learned}, state-space models~\cite{qin2024mambavc} and invertible networks~\cite{ma2020end}. Subsequently, the continuous transformed coefficients are quantized into the discrete representation, wherein the uniform scalar quantization~\cite{balle2016end} is commonly adopted. As vector quantization can, in principle, better approach the rate-distortion bound (Eqn.(\ref{eq:RD_Function})), low-complexity vector quantization~\cite{agustsson2017soft,zhang2023lvqac,feng2023nvtc} and dependent scalar quantization~\cite{suhring2022trellis,ge2024nlic} schemes are also studied. Finally, once the quantized latent representations are obtained, a neural entropy model is essential for performing entropy coding. Herein, since the latent cannot be perfectly decorrelated, an accurate entropy model that can better capture the dependencies and closely match the ground-truth latent distribution is crucial for the overall network performance. Prominent research directions such as spatial-channel context mining~\cite{minnen2018joint,minnen2020channel,guo2021causal,jiang2023mlic}, expressive latent distribution modeling~\cite{cheng2020learned,zhu2022unified,fu2023learned} and latent architecture designs~\cite{balle2018variational,he2021checkerboard,hu2020coarse} are explored.

\paragraph{Optimization over quantized latent representations.} The surrogate-based methods~\cite{balle2016end,agustsson2017soft,yang2020improving,agustsson2020universally,zhang2023uniform} have been shown to effectively address the non-differentiable quantization function during end-to-end optimization. In addition to AUN~\cite{balle2016end}, various quantization surrogates such as soft assignment~\cite{agustsson2017soft}, Gumbel-softmax trick~\cite{yang2020improving}, and soft rounding function~\cite{agustsson2020universally} have been studied. However, as the surrogate introduces a train-test mismatch in forward calculations, the annealing techniques~\cite{agustsson2017soft,yang2020improving,agustsson2020universally,zhang2023uniform} are then proposed to control the ``softness'' of surrogates, gradually approximating the hard quantization function as training progresses. The straight-through estimator (STE)~\cite{theis2017lossy} provides an alternative approach to address the train-test mismatch by enforcing an identity mapping during the backward calculation, thereby enabling direct hard quantization in the forward pass. Nevertheless, STE ultimately introduces a significant gradient estimation bias for the analysis transform. To counter this, an additional fine-tuning stage for the entropy model and synthesis transform has been suggested~\cite{guo2021soft}. 

While numerous network optimization methods have been developed from a quantization perspective, the regularization view has largely escaped research attention. To the best of our knowledge, the most related efforts are regularization methods for variational autoencoders (VAEs), such as regularization on latent priors~\cite{zhao2018adversarially,xu2020learning,wu2020vector}, reconstruction~\cite{le2018supervised,ma2018constrained}, and consistency~\cite{sinha2021consistency}. However, as VAEs are designed for generative modeling, these methods are not applicable to optimizing compression networks. To bridge this knowledge gap, this paper proposes a structural regularizer for neural image compression and conducts extensive experiments to demonstrate its superiority.




%% file: sec/3_method.tex
\section{Methodology}
In this section, we begin by deriving the equivalent maximization form of the latent entropy minimization problem for both the direct and transform coding models. Subsequently, building on this derivation, we propose a novel regularization method for optimizing the neural compression networks. Note that during our derivations, we use the naive transform coding model without the side information branch (i.e., the $\bm{\hat{Z}}$ depicted in Fig~\ref{fig.compression_model}(c)). This significantly simplifies our analysis without compromising much practicality, as it has been shown that the information flowing into the side branch is negligible compared to the latent branch~\cite{balle2018variational}.

\subsection{Motivation: from minimization to maximization}\label{sec:motivation}
\begin{Lemma}\label{lemma1:diret-eqn}
Considering a deterministic quantization process ${Q}(\cdot)$ and a deterministic dequantization process ${Q}^{-1}(\cdot)$ for the direct coding model illustrated in \cref{fig.compression_model}(a), the following equalities hold automatically:
\begin{numcases}{}
	I(\bm{X};\bm{\hat{X}})=H(\bm{\hat{X}}) \label{eq:diret-eq1} \\
        H(\bm{\hat{X}})=H(\bm{U}) \label{eq:diret-eq2} \\
        H(\bm{U})=I(\bm{X};\bm{\hat{X}}) \label{eq:diret-eq3}
\end{numcases}
\end{Lemma}
\begin{proof}\renewcommand{\qedsymbol}{}
Eqn.(\ref{eq:diret-eq1}) follows the facts that $H(\bm{\hat{X}}|\bm{X})=0$, meaning there is no uncertainty in $\bm{\hat{X}}$ given the observation of $\bm{X}$, and $I(\bm{X};\bm{\hat{X}})=H(\bm{\hat{X}})-H(\bm{\hat{X}}|\bm{X})$. Similarly, Eqn.(\ref{eq:diret-eq2}) follows the fact that $H(\bm{U}|\bm{\hat{X}})=H(\bm{\hat{X}}|\bm{U})=0$ and $H(\bm{\hat{X}})+H(\bm{U}|\bm{\hat{X}})=H(\bm{U})+H(\bm{\hat{X}}|\bm{U})$. Eqn.(\ref{eq:diret-eq3}) is immediately apparent from Eqn.(\ref{eq:diret-eq1}) and Eqn.(\ref{eq:diret-eq2}). For more details, readers may refer to the Supplementary Material (\cref{sec:Proof-of-Lemma1}).
\end{proof}

\begin{Theorem}\label{theo:diret-minmax}
For the direct coding model elaborated in Lemma~\ref{lemma1:diret-eqn}, we have
\begin{gather}
    H(\bm{U}) = H(\bm{X})-H(\bm{X}|\bm{\hat{X}}). \label{eq:diret-eq5}
\end{gather}
The informational diagram is thus illustrated as Fig.\ref{fig.compression_model}(b).
\end{Theorem}
\begin{proof}\renewcommand{\qedsymbol}{}
This is immediately apparent from Eqn.(\ref{eq:diret-eq3}), as the right-hand-side of Eqn.(\ref{eq:diret-eq5}) is by definition $I(\bm{X};\bm{\hat{X}})$.
\end{proof}

From Theorem~\ref{theo:diret-minmax}, for the direct coding model, it is clear that minimizing the latent entropy $H(\bm{U})$ can be equivalently achieved by maximizing the conditional source entropy $H(\bm{X}|\bm{\hat{X}})$, for any known source $\bm{X}$. In practice, this aligns with the training setting of the neural compression network, wherein the training set is given in advance, and thus $H(\bm{X})$ is fixed. However, as the prevailing neural compression networks are based on the transform coding model, in the following discussion, we extend this conclusion to the transform coding setting.

\begin{Lemma}\label{lemma:trans-eqn}
Considering a transform coding model built on the direct coding model elaborated in Lemma~\ref{lemma1:diret-eqn} with a deterministic analysis transform $T_A$ and a deterministic synthesis transform $T_S$ (\cref{fig.compression_model}(c)), the following equalities hold automatically:
\begin{numcases}{}
	I(\bm{X};\bm{\hat{X}})=H(\bm{\hat{X}}) \label{eq:trans-eq1} \\
        H(\bm{\hat{X}})=H(\bm{U})-H(\bm{U}|\bm{\hat{X}}) \label{eq:trans-eq2} \\
        H(\bm{U})=I(\bm{X};\bm{\hat{X}})+H(\bm{U}|\bm{\hat{X}}) \label{eq:trans-eq3}
\end{numcases}
\end{Lemma}
\begin{proof}\renewcommand{\qedsymbol}{}
Compared to the proof of Lemma~\ref{lemma1:diret-eqn}, for the transform coding, we still have $H(\bm{\hat{X}}|\bm{X})=0$ and $H(\bm{\hat{X}}|\bm{U})=0$. However, since the synthesis transform $T_S$ is not guaranteed to be a bijection mapping, the $H(\bm{U}|\bm{\hat{X}})=0$ does not always hold, leading to the correction of Eqn.(\ref{eq:trans-eq2}) and thus Eqn.(\ref{eq:trans-eq3}). For more details, readers may refer to the Supplementary Material (\cref{sec:Proof-of-Lemma2}).
\end{proof}
\begin{Theorem}\label{theo:trans-minmax}
For the transform coding model elaborated in Lemma~\ref{lemma:trans-eqn}, we have
\begin{gather}
    H(\bm{U}) = H(\bm{X})-H(\bm{X}|\bm{\hat{X}})+H(\bm{U}|\bm{\hat{X}}).\label{eq:trans-eq4}
\end{gather}
\end{Theorem}
\begin{proof}\renewcommand{\qedsymbol}{}
This is immediately apparent from Eqn.(\ref{eq:trans-eq3}).
\end{proof}
\paragraph{Remarks.} Two conclusions can be drawn for the transform coding model.
\begin{enumerate}[label=(\alph*)]
\item \textbf{Equivalence of minimization and maximization.} From Theorem~\ref{theo:trans-minmax}, for any known source $\bm{X}$, minimizing $H(\bm{U})$ can be equivalently achieved by maximizing the $H(\bm{X}|\bm{\hat{X}})-H(\bm{U}|\bm{\hat{X}})$. For a bijection synthesis function $T_S$, of which $H(\bm{U}|\bm{\hat{X}})$ is zero, this maximization objective is reduced to $H(\bm{X}|\bm{\hat{X}})$, which is the same as the direct coding;
\item \textbf{Achieving $R(D)$ by minimizing $H(\bm{U})$.} From Eqn.(\ref{eq:trans-eq3}), minimizing $H(\bm{U})$ is simultaneously minimizing $I(\bm{X};\bm{\hat{X}})$ and $H(\bm{U}|\bm{\hat{X}})$. In the ideal case, the informational lower bound given in Eqn.(\ref{eq:RD_Function}) is achieved with $H(\bm{U}|\bm{\hat{X}})$ being reduced to 0. 
\end{enumerate}

\begin{figure}[h]
\centering
\includegraphics[width=0.42\textwidth]{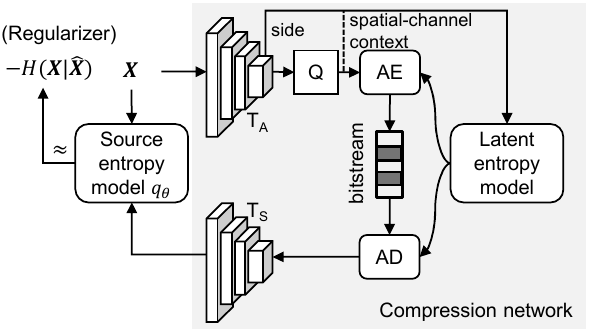}
\caption{An illustration of the proposed regularization method, wherein an additional source entropy model $q_\theta$ is introduced. The $\mathrm{AE}$ and $\mathrm{AD}$ represent arithmetic encoding and arithmetic decoding, respectively. The side information branch (i.e., the $\bm{\hat{Z}}$ branch) is considered as part of the latent entropy model, which necessitates bits transmission.}
\label{fig.system}
\end{figure}
\begin{algorithm}
\caption{Training Strategy}\label{alg:training}
\begin{algorithmic}[1]
\For{each training step}
    \State Sample a minibatch of training images. 
    \State Perform the forward pass of the compression network and source entropy model $q_{\theta}$.
    \State Compute the backward gradient and update the compression network based on Eqn.(\ref{eq:proposed-loss}).
    \State Freeze the compression network.
    \State Perform the forward pass of $q_{\theta}$.
    \State Compute the backward gradient and update the source entropy model $q_{\theta}$ based on Eqn.(\ref{eq:source-entropy}).
\EndFor
\end{algorithmic}
\end{algorithm}
\subsection{Information-theoretic regularizer}\label{sec:regularizer}
In principle, optimizing neural compression networks with a quantization surrogate, e.g., AUN, introduces gradient estimation errors~\cite{zhang2023uniform}, thereby deviating the objective of minimizing the latent entropy $H(\bm{U})$ under certain distortion constraints (Eqn.(\ref{eq:rate-dist-loss})), harming both the efficiency and efficacy of the optimization. Motivated by the equivalence of the minimization and maximization optimization discussed in~\cref{sec:motivation}, we propose to address this issue by imposing structural maximization constraints on the optimization process. As such, the vanilla objective of minimizing $H(\bm{U})$ can be further enhanced through this maximization regularization.

From Theorem~\ref{theo:trans-minmax}, an ideal gradient direction for model updating is one that simultaneously maximizes $H(\bm{X}|\bm{\hat{X}})-H(\bm{U}|\bm{\hat{X}})$,
which can be further interpreted as maximizing $H(\bm{X}|\bm{\hat{X}})$ while minimizing $H(\bm{U}|\bm{\hat{X}})$. To integrate this regularization into the training objective, two parameterized distributions $q_{\theta}(\bm{X}|\bm{\hat{X}})$ and $q_{\varphi}(\bm{U}|\bm{\hat{X}})$ are necessitated.
However, accurately modeling an entropy model $q_{\varphi}(\bm{U}|\bm{\hat{X}})$ alongside the compression network is quite challenging, as it involves learning another ``encoder'' function from the reconstruction $\bm{\hat{X}}$ to the latent $\bm{U}$. We experimentally find that a poorly learned $H(\bm{U}|\bm{\hat{X}})$ regularizer may harm the optimization performance, and that the increased training complexity is also burdensome. To balance practical feasibility with regularization accuracy, we include only the first $H(\bm{X}|\bm{\hat{X}})$ term for regularization, which is an upper bound for the ground-truth $H(\bm{X}|\bm{\hat{X}})-H(\bm{U}|\bm{\hat{X}})$ objective. The entropy modeling becomes more tractable since the reconstruction $\bm{\hat{X}}$ is closely aligned with the source $\bm{X}$. The primary rationale for such simplification is that although maximizing $H(\bm{X}|\bm{\hat{X}})$ alone is less precise than $H(\bm{X}|\bm{\hat{X}})-H(\bm{U}|\bm{\hat{X}})$, it still provides a meaningful constraint on the ”noisy” gradient, yielding better gradient direction than no regularization. Another justification is that since we will retain the $H(\bm{U})$ minimization term in the overall objective, this $H(\bm{U}|\bm{\hat{X}})$ gap, in principle (Eqn.(\ref{eq:trans-eq4})), should decrease as training progresses, which ensures that the regularization effect remains aligned with the ideal maximization objective. The resulting training objective is thus given by:
\begin{align}\label{eq:proposed-loss}
\mathop{\text{min}}\ R+\lambda D+\alpha\underbrace{\mathbb{E}_{\bm{X}}[\text{log}\ q_{\theta}(\bm{X}|\bm{\hat{X}})]}_{\approx-H(\bm{X}|\bm{\hat{X}})},
\end{align}
where $R+\lambda D$ follows the rate-distortion loss in Eqn.(\ref{eq:rate-dist-loss}); $-H(\bm{X}|\bm{\hat{X}})$ is the proposed regularizer; $q_{\theta}(\bm{X}|\bm{\hat{X}})$ is the estimated source distribution; $\alpha$ is regularization factor, controlling the degree of regularization. An illustration for this regularization method is given in Fig.(\ref{fig.system}). In addition to the compression network, $q_{\theta}(\bm{X}|\bm{\hat{X}})$ is introduced, which is parameterized by a neural network and can be optimized via the maximum likelihood criteria~\cite{goodfellow2016deep}: 
\begin{align}\label{eq:source-entropy}
\mathop{\text{max}}\ \mathbb{E}_{\bm{X}}[\text{log}\ q_{\theta}(\bm{X}|\bm{\hat{X}})].
\end{align}
This results in a max-min optimization problem where minimizing Eqn.(\ref{eq:proposed-loss}) simultaneously creates a conflict. To resolve this, a GAN-style two-stage joint training strategy~\cite{goodfellow2014generative} is employed: At each training step, we first update the compression network using Eqn.(\ref{eq:proposed-loss}), then freeze it and update $q_{\theta}(\bm{X}|\bm{\hat{X}})$ using Eqn.(\ref{eq:source-entropy}). We summarize the strategy in \cref{alg:training}.

%% file: sec/4_experiment.tex
\begin{figure*}[h]
\centering
\includegraphics[width=0.98\textwidth]{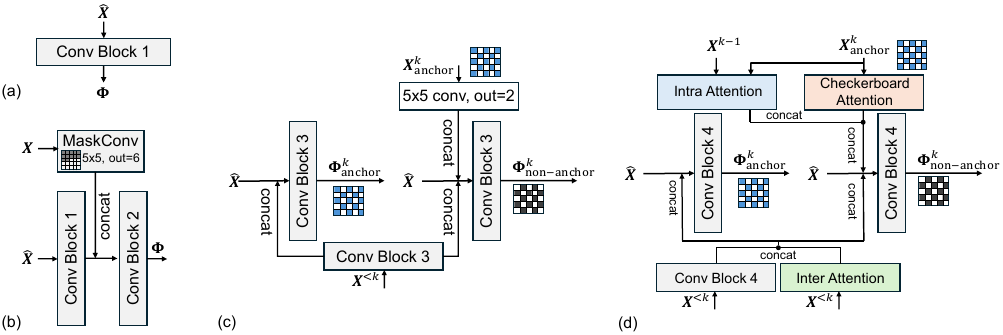}
\caption{Conditional source entropy modeling for (a) \textit{hyperprior}~\cite{balle2018variational}, (b) \textit{autoregressive}~\cite{minnen2018joint} and \textit{attention}~\cite{cheng2020learned}, (c) EILC~\cite{he2022elic}, and (d) MLIC++~\cite{jiang2023mlic}. The architecture and module designs are aligned with the latent entropy model. Additional details regarding the specific modules are provided in the Supplementary Material (\cref{sec:source-model}).}
\label{fig.cond-source-model-main}
\end{figure*}

\section{Experiments}\label{sec:ex}
\subsection{Experimental setup}
\paragraph{Compression models.}Three classic image compression models are used in exhaustive validation experiments: the hyperprior context model~\cite{balle2018variational} (with scale and mean~\cite{minnen2018joint}), the joint hyperprior and spatial autoregressive context model~\cite{minnen2018joint}, and the attention-based transform model~\cite{cheng2020learned}. We may refer to these models as \textit{hyperprior}, \textit{autoregressive}, and \textit{attention}, respectively. Moreover, we conduct the experiments on two advanced compression models: ELIC~\cite{he2022elic} and MLIC++~\cite{jiang2023mlic}. Notably, ELIC utilizes checkerboard and channel-conditioned latent entropy contexts, while MLIC++ builds upon ELIC by introducing additional attention-based context designs and is identified as one of the state-of-the-art (SOTA) models. 

We train these models from scratch using the vanilla rate-distortion loss (Eqn.(\ref{eq:rate-dist-loss})), and compare their performance when trained with the proposed regularization method (\cref{alg:training}) under an equal number of training steps. For the implementations of the \textit{hyperprior}, \textit{autoregressive}, and \textit{attention} models, we adopt CompressAI's re-implementation~\cite{begaint2020compressai}, where the quantization surrogate is ANU. For ELIC and MLIC++, we use the open re-implementations from~\cite{jiang2022unofficialelic} and the official implementation, respectively, both of which employ a mixed ANU-STE quantization surrogate~\cite{minnen2020channel}.



\begin{figure*}[h]
\centering
\includegraphics[width=0.95\textwidth]{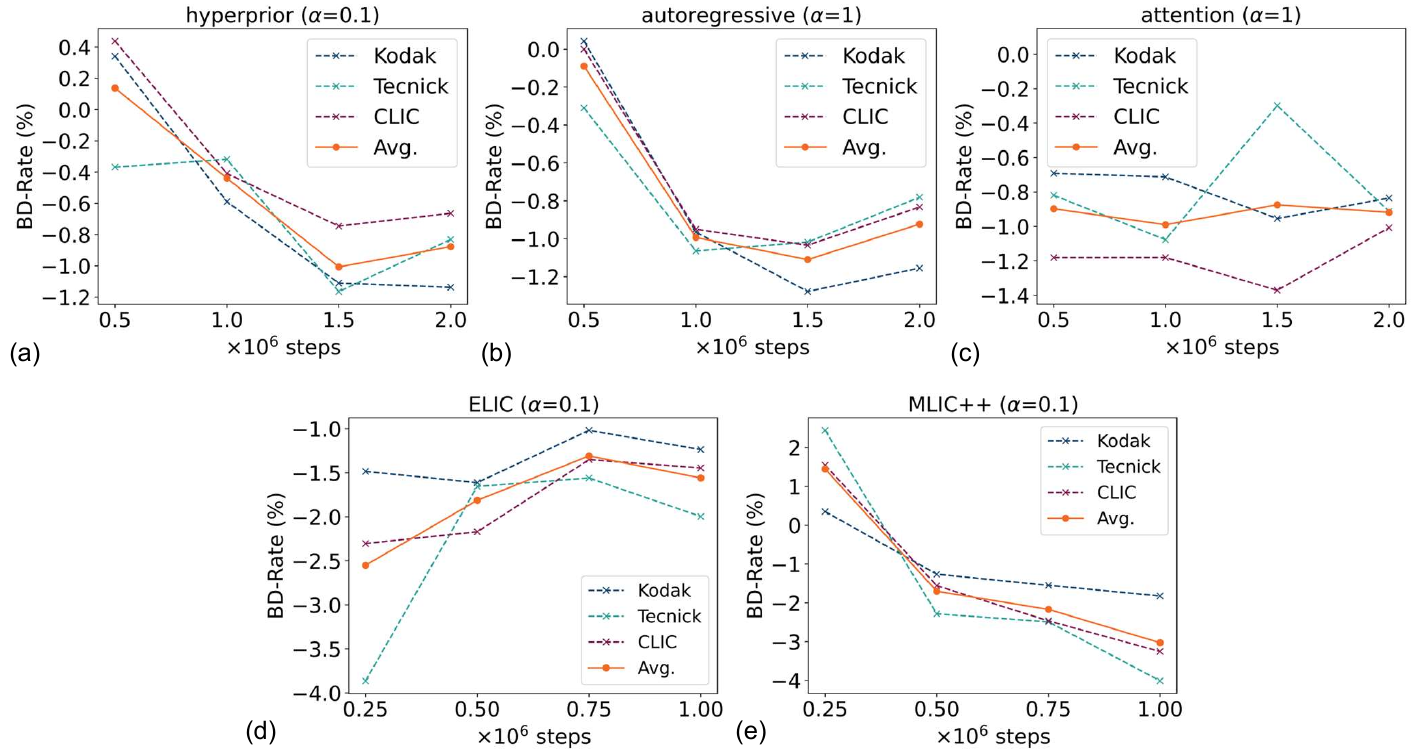}
\caption{Performance of the proposed regularization method on the \textit{hyperprior}~\cite{balle2018variational}, \textit{autoregressive}~\cite{minnen2018joint}, \textit{attention}~\cite{cheng2020learned}, ELIC~\cite{he2022elic}, and MLIC++~\cite{jiang2023mlic}. The anchor is trained with the vanilla rate-distortion loss (Eqn.(\ref{eq:rate-dist-loss})) under an equal number of training steps. $\alpha$ indicates the regularization factor (Eqn.(\ref{eq:proposed-loss})).}
\label{fig.overall-performance}
\end{figure*}

\paragraph{Conditional source entropy modeling.} For each compression model, we apply exactly the same entropy modeling architecture as the latent entropy modeling $q_{\phi}(\bm{U}|\bm{\hat{Z}})$. Our experiments indicate that aligning the entropy model between the latent and source is critical for effective regularization; deviations in either direction (stronger or weaker source models) degrade compression performance.
For the \textit{hyperprior} model~\cite{balle2018variational}, $q_{\theta}(\bm{X}|\bm{\hat{X}})$ is modeled as the factorized Gaussian distribution conditioned on $\bm{\hat{X}}$, i.e.,
\begin{gather}
     q_\theta(\bm{X}|\bm{\hat{X}}) \sim \prod \limits_{i=1}^N\mathbb{N}(\mu_i,\sigma_i^2), \\
     \text{where} \ \bm{\Phi}\triangleq(\mu_1,...,\mu_N,\sigma_1^2,...,\sigma_N^2)=f_\theta(\bm{\hat{X}}),
\end{gather}
and $f_\theta$ is modeled by convolutional layers; $\mathbb{N}$ denotes the Gaussian distribution; $\bm{\Phi}$ denote the mean and variance; $N$ denotes the dimension of both $\bm{X}$ and $\bm{\hat{X}}$. For both the \textit{autoregressive} and \textit{attention} models, the mean and variance are determined by the additional spatial context, i.e.,
\begin{gather} (\mu_1,...,\mu_i,\sigma_1^2,...,\sigma_i^2)=f_\theta(\bm{\hat{X}},\bm{X}_{<i}),
\end{gather}
where $\bm{X}_{<i}$ denotes the pixels before the current position and is captured by a $5\times5$ masked convolution layer. For ELIC, $\bm{\Phi}$ is divided by a checkerboard pattern, leading to the anchor part $\bm{\Phi}_{\text{anchor}}$ and the non-anchor part $\bm{\Phi}_{\text{non-anchor}}$. Both the original frame $\bm{X}$ and $\bm{\Phi}$ are split into 3 slices for context modeling. In the first slice, $\bm{\Phi}_{\text{anchor}}^1$ is directly determined by $\bm{\hat{X}}$, while $\bm{\Phi}_{\text{non-anchor}}^1$ is determined by an additional spatial context from $\bm{X}^1_{\text{anchor}}$. For $k^{th}\ (k>1)$ slice, both the $\bm{\Phi}_{\text{anchor}}^k$ and $\bm{\Phi}_{\text{non-anchor}}^k$ are conditioned on extra channel context from previous slices $\bm{X}^{<k}$. For MLIC++, following the approach in latent entropy modeling~\cite{jiang2023mlic}, the checkerboard attention module is employed to capture the local dependencies within each slice. Additionally, the intra attention module, which takes the original anchor slice $\bm{X}^k_{\text{anchor}}$ and the previous slice $\bm{X}^{k-1}$ as input, captures global dependencies within slices. Furthermore, the inter attention module is introduced to capture the global dependency of inter slices. The detailed network structure of $q_\theta(\bm{X}|\bm{\hat{X}})$ for different models is illustrated in Fig.~\ref{fig.cond-source-model-main}. For further details on the convolutional modules, intra-attention module, inter-attention module, and checkerboard attention module, readers may refer to the Supplementary Material (\cref{sec:source-model}).


\paragraph{Training.}For the training set, we use Flickr20k~\cite{liu2020unified}, consisting of 20745 natural images from the~\textit{Flickr.com}. The train-to-validation ratio is set to 9:1. To eliminate the effect of randomness and facilitate a fair performance comparison, the random seed is fixed to 1 for all training. Following the settings of CompressAI, we train four bit-rate points for each model, i.e., $\lambda\in\{0.0018, 0.0035, 0.0067, 0.0130\}$. The image patch size is set to $256\times256$. For the classic \textit{hyperprior}, \textit{autoregressive}, and \textit{attention} models, the batch size is set to 16. The Adam optimizer~\cite{kingma2014adam} is employed, with learning rates of $10^{-4}$ for the compression model and $10^{-3}$ for the source entropy model. The training proceeds $2\times10^6$ steps. For the advanced MLIC++ model, the learning rate of the source entropy model is reduced to $10^{-4}$, with a batch size of 8 and training duration of $1\times10^6$ steps. The training setting of the ELIC model is aligned with MLIC++. All experiments are conducted on Nvidia RTX 4090 GPUs.


\paragraph{Performance evaluation.}We follow the convention of evaluating over natural image datasets of Kodak~\cite{kodak1993kodak}, CLIC 2024 Validation Set~\cite{CLIC} and Tecnick~\cite{asuni2014testimages}. Besides, to demonstrate generalization performance, we follow the domains in~\cite{zhang2024few} and evaluate the models on four out-of-domain datasets\footnote{We exclude the crater dataset due to insufficient documentation on its source and collection methodology.}: (a) pixel-style dataset~\cite{lv2023dynamic} (100 images in total); (b) screen-content dataset SCI1K~\cite{yang2021implicit} (the first 100 images of the training set); (c) game dataset CCT-CGI~\cite{min2017unified} (24 images in total); (d) pathology dataset BRACS~\cite{brancati2022bracs} (81 images in total\footnote{Folder: BRACS\_RoI$\backslash$latest\_version$\backslash$test$\backslash$0\_N}).


\subsection{Results}
\paragraph{Performance on natural images. }In Fig.\ref{fig.overall-performance}, we present the BD-Rate comparison at four different training steps. The anchor is trained with the vanilla rate-distortion loss (Eqn.(\ref{eq:rate-dist-loss})) under an equal number of training steps. The regularization factors $\alpha$ for the \textit{hyperprior}, \textit{autoregressive}, \textit{attention}, ELIC, and MLIC++ models are set to 0.1, 1, 1, 0.1, and 0.1, respectively. It can be observed that, for training steps beyond $0.5\times10^{6}$ across all five models, our proposed regularization method can achieve better compression performance than the vanilla optimization method. In particular, at the $2\times10^{6}$ step, the average BD-Rates for \textit{hyperprior}, \textit{autoregressive}, and \textit{attention} are -0.88\%, -1.06\% and -0.92\%, respectively; at the $1\times10^{6}$ step, the average BD-Rates for ELIC and MLIC++ are -1.56\% and -3.03\%, respectively. Meanwhile, the results demonstrate that the regularization effect varies across models as training progresses: For the \textit{hyperprior}, \textit{autoregressive}, and MLIC++ models, after a brief convergence period, the BD-Rates are gradually improved from 0 to around -1\%. For the ELIC model, the regularization is most effective during the earlier training stages, achieving over 2.5\% BD-Rates saving at the $0.25\times10^{6}$ step. For the \textit{attention} model, the regularization achieves a relatively stable coding gain.

\begin{figure}[h]
\centering
\includegraphics[width=0.38\textwidth]{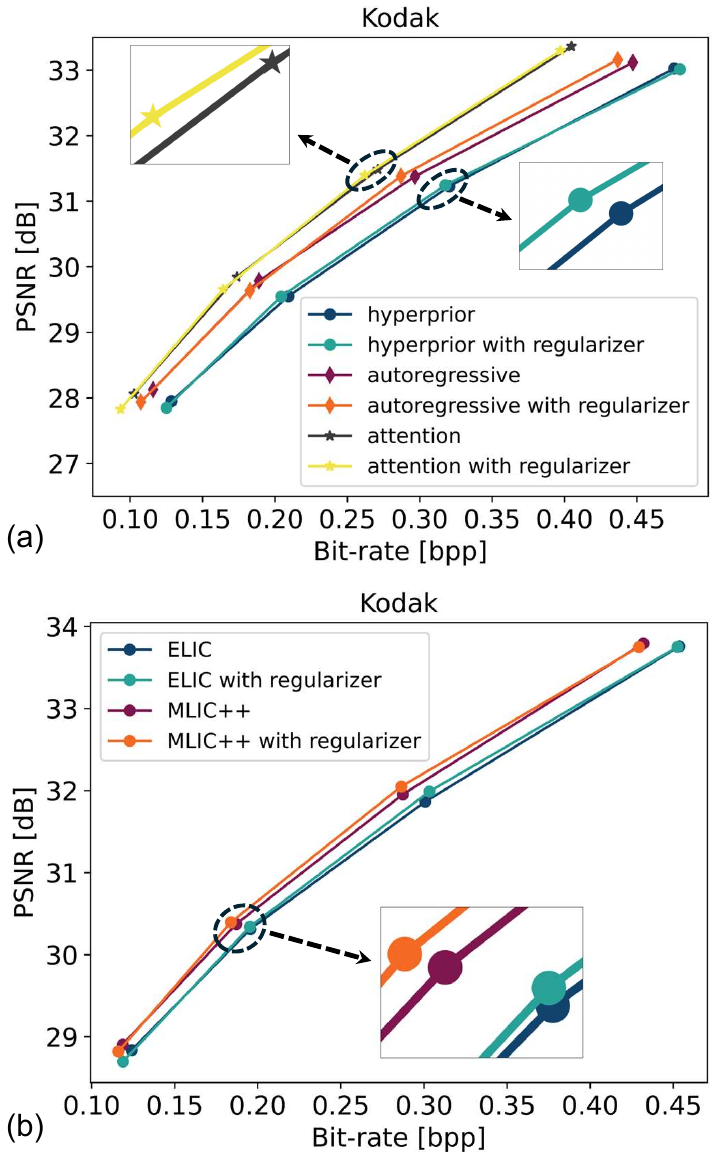}
\caption{Compression performance on Kodak. Our regularizer achieves -1.13\%, -1.57\%, -0.84\%, -1.24\%, and -1.82\% BD-Rates for the \textit{hyperprior}, \textit{autoregressive}, \textit{attention} models, ELIC, and MLIC++, respectively.}
\label{fig.200Wstep-performance-Kodak}
\end{figure}

In Fig.~\ref{fig.200Wstep-performance-Kodak}, we present the detailed rate-distortion (RD) curves on the Kodak dataset. For the \textit{hyperprior}, \textit{autoregressive}, and \textit{attention} models, the results correspond to the $2\times10^{6}$ training step, while for the ELIC and MLIC++ models, the results are from the  $1\times10^{6}$ step. The results indicate that the introduction of the additional conditional source entropy regularization leads to a slight reduction in the bit rate for most of the points, although a few exceptions exhibit an increase due to training noise. From the form of regularization in Eqn.(\ref{eq:proposed-loss}), i.e., $-\alpha H(\bm{X}|\bm{\hat{X}})$, the extent of such bit rate reduction depends on both the regularization factor $\alpha$ and the estimated conditional source entropy $H(\bm{X}|\bm{\hat{X}})$. 

\tabcolsep=0.08cm
\begin{table}[htbp]\footnotesize
  \centering
  \caption{Generalization performance of the proposed regularization method on four out-of-domain datasets, i.e., the pixel-style~\cite{lv2023dynamic}, screen~\cite{yang2021implicit}, game~\cite{min2017unified}, and pathology~\cite{brancati2022bracs} datasets. The in-domain performance on natural images~\cite{kodak1993kodak,CLIC,asuni2014testimages} is also listed for reference.}
    \begin{tabular}{cccccc}
    \toprule
    \multicolumn{1}{c}{\multirow{2}[4]{*}{Test set}} & \multicolumn{5}{c}{Regularized v.s. Unregularized} \\
\cmidrule{2-6}          & \textit{hyperprior} & \textit{autoregressive} & \multicolumn{1}{l}{\textit{attention}} & ELIC  & MLIC++ \\
    \midrule
    Kodak & -1.14\% & -1.57\% & -0.84\% & -1.24\% & -1.82\% \\
    \midrule
    Tecnick & -0.83\% & -0.78\% & -0.91\% & -2.00\% & -4.01\% \\
    \midrule
    CLIC  & -0.66\% & -0.83\% & -1.01\% & -1.45\% & -3.25\% \\
    \midrule
    \textbf{Natural Avg.} & \textbf{-0.88\%} & \textbf{-1.06\%} & \textbf{-0.92\%} & \textbf{-1.56\%} & \textbf{-3.03\%} \\
    \midrule
    Pixel-style & -2.20\% & -1.96\% & -0.47\% & -2.71\% & -3.08\% \\
    \midrule
    Screen & -1.15\% & -2.01\% & -0.69\% & -1.02\% & -1.72\% \\
    \midrule
    Game  & -1.31\% & -0.83\% & -0.82\% & -2.04\% & -3.44\% \\
    \midrule
    Pathology & -1.11\% & -0.81\% & -2.38\% & -1.82\% & -0.43\% \\
    \bottomrule
    \end{tabular}%
  \label{tab:generalization}%
\end{table}%

\paragraph{Generalizing on unseen domains.} In machine learning, regularization plays an essential role in preventing overfitting and enhancing generalization. In Table~\ref{tab:generalization}, we directly apply the models trained on natural images to the four unseen domains and evaluate the effectiveness of our regularization method. The results indicate that, across all new domains, regularized models consistently outperform unregularized ones. Performance improvements vary by domain and model, with notable gains in some cases. For instance, in the pathology domain with the \textit{attention} model, the regularization achieves a BD-Rate of -2.38\%, more than double the improvement observed in the natural domain.

\begin{figure*}[h]
\centering
\includegraphics[width=0.98\textwidth]{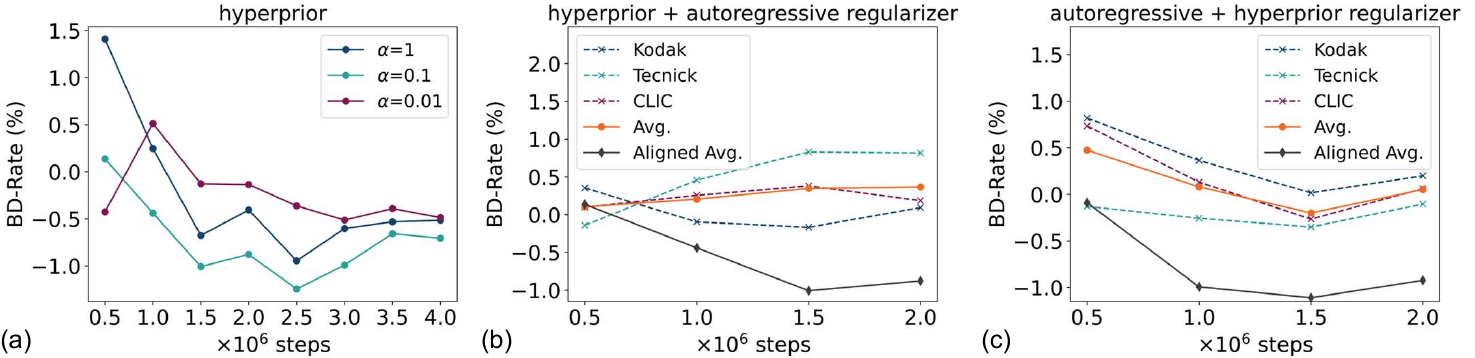}
\caption{(a) Effects of regularization factor $\alpha$. (b) Regularization performance for the \textit{hyperprior} compression model with the \textit{autoregressive} regularizer; (b) the \textit{autoregressive} compression model with the \textit{hyperprior} regularizer.}
\label{fig.alpha-400Wstep}
\end{figure*}

\paragraph{Effect of $\alpha$ and longer training.} In Fig.(\ref{fig.alpha-400Wstep})(a), we continue training both the unregularized and regularized \textit{hyperprior} models to $4\times10^{6}$ steps and demonstrate the effect of the regularization factor $\alpha$ on the performance. Herein, with extended training, the regularization consistently enhances optimization. The best performance, with $\alpha=1$, is achieved at the $2.5\times10^{6}$ step, showing a BD-Rate of -1.24\%. As training progresses to $4\times10^{6}$ steps, the gain reduces to around 0.7\%. For the effect of hyperparameter $\alpha$, it is shown that either the value is too large or too small will not benefit the optimization, which can be explained as follows. As discussed in~\cref{sec:regularizer}, our $\alpha H(\bm{X}|\bm{\hat{X}})$ maximization regularization is essentially a bound upper of the ground-truth objective $\alpha(H(\bm{X}|\bm{\hat{X}})-H(\bm{U}|\bm{\hat{X}}))$. When $\alpha$ is large, this $H(\bm{U}|\bm{\hat{X}})$ gap is also amplified, which harms the regularization accuracy and thus the optimization performance. When $\alpha$ is small, the regularizer has little contribution to the overall loss, and the optimization is reduced to the vanilla objective. 

\paragraph{On the alignment of the entropy models. }In our experimental setup, we emphasize the advantages of aligned entropy modeling between latent and source distributions. Herein, we extend the analysis to explore cases where the entropy models are unaligned, investigating two scenarios: a compression model equipped with (a) a stronger source entropy model and (b) a weaker source entropy model. In particular, for the source models of the \textit{hyperprior} and \textit{autoregressive} depicted in Fig.\ref{fig.cond-source-model-main}(a) and Fig.\ref{fig.cond-source-model-main}(b), respectively, we consider the \textit{autoregressive} entropy model with additional spatial contexts to be a more accurate and thus ``stronger'' model. Thereafter, we train the \textit{hyperprior} compression model with a \textit{autoregressive} source regularizer and the \textit{autoregressive} compression model with a \textit{hyperprior} source regularizer, and the results are depicted in Fig.\ref{fig.alpha-400Wstep}(b) and Fig.\ref{fig.alpha-400Wstep}(c), respectively. The values of $\alpha$ are still set to 0.1 and 1 for the \textit{hyperprior} and \textit{autoregressive} regularizers, respectively. The performance of the aligned entropy model is also included for reference. From Fig.\ref{fig.alpha-400Wstep}, it can be observed that neither a stronger nor a weaker source entropy model consistently facilitates the optimization performance of the compression models. Specifically, when a stronger \textit{autoregressive} regularizer is imposed on the \textit{hyperprior} network, the average BD-Rate remains positive throughout training. When a weaker \textit{hyperprior} regularizer is employed by the \textit{autoregressive} network, the optimization is temporarily improved at the $1.5\times10^{6}$ training step, with a BD-Rate of -0.20\%. In contrast, aligning the entropy models leads to significant performance improvements.




\subsection{Limitations}
Our regularization method is applied exclusively during the network training stage and therefore introduces no additional inference complexity. However, since it involves training an additional source entropy model $q_{\theta}(\bm{X}|\bm{\hat{X}})$, the training complexity is inevitably increased. A straightforward yet meaningful analysis can be provided as follows. In our experiments, we demonstrate that the best regularization performance is achieved when the neural architectures of the source and latent entropy models are aligned. Assuming the complexities of the source and latent models are approximately equal, the complexity ratio can be calculated as:
\begin{equation}\label{eq:complexity}
    \frac{2C_L+C_T}{C_L+C_T} = 1 + \frac{1}{1+C_T/C_L}, 
\end{equation}
where $C_L$ and $C_T$ denote the complexity of the latent entropy model and the transform backbone, respectively. From Eqn.(\ref{eq:complexity}), it is indicated that the complexity is highly related to $C_T/C_L$. When the transform backbone is significantly more complex than the latent entropy model, i.e., $C_T/C_L\to \infty$, the overhead becomes less prominent. For $\lambda=0.0018$, we train $1\times10^5$ steps for each model on the same machine (Nvidia RTX 4090) and compute the time ratio. As shown in Table.~\ref{tab:complexity}, compared to the unregularized baseline, the training time of our proposed method increases by 28.8\%, 18.2\%, 8.1\%, 39.3\% and 29.8\% for the \textit{hyperprior}, \textit{autoregressive}, \textit{attention}, ELIC and MLIC++ models, respectively. Notably, for the \textit{attention} model, the overhead is the least significant. This can be attributed to the fact that the attention-based transform backbone is significantly more complex than the entropy model, as illustrated in Fig.~\ref{fig.cond-source-model-main}(b). To further reduce training complexity, one can employ lightweight convolutional layers, such as depthwise separable convolution~\cite{chollet2017xception}, or improve the iterative training strategy outlined in Algorithm~\ref{alg:training}.


\tabcolsep=0.08cm
\begin{table}[htbp]\footnotesize
  \centering
  \caption{Training complexity of the proposed regularizer compared to the unregularized anchor.}
    \begin{tabular}{cccccc}
    \toprule
          & \textit{hyperprior} & \textit{autoregressive} & \textit{attention} & ELIC  & MLIC++ \\
    \midrule
    Training time & 128.8\% & 118.2\% & 108.1\% & 139.3\% & 129.8\% \\
    \bottomrule
    \end{tabular}%
  \label{tab:complexity}%
\end{table}%

%% file: sec/5_conclusion.tex
\section{Conclusion}
This paper investigates a structural entropy regularizer for lossy neural image compression. Specifically, motivated by the information-theoretic analysis of compression models, we demonstrate that, in addition to minimizing latent entropy, maximizing conditional source entropy can be equally important. Subsequently, by regularizing the neural networks to simultaneously maximize the conditional source entropy during end-to-end optimization, we experimentally observe improvements in both the in-domain compact representation and the out-of-domain generalization abilities of compression models. Additionally, as the primary overhead of this regularizer comes from training an additional source entropy model, the proposed method imposes no added inference complexity, 
opening up a new space for regularizing neural compression networks.


%% file: sec/X_suppl.tex
\clearpage
\setcounter{page}{1}
\maketitlesupplementary

\section{Proof of Lemma~\ref{lemma1:diret-eqn}}\label{sec:Proof-of-Lemma1}
For a deterministic quantization process ${\rm Q}(\cdot)$, the conditional probability $p(\bm{U}|\bm{X})$ can only take values of 0 or 1, i.e.,
\begin{equation}
p(\bm{U}|\bm{X})=
\left\{
     \begin{array}{lr}
     1, \quad \text{if}\ \ \bm{U}={\rm Q}(\bm{X})  \\
     0, \quad \text{if}\ \ \bm{U}\neq{\rm Q}(\bm{X}) 
     \end{array}
\right..
\end{equation}
Similarly, for a deterministic dequantization process ${\rm Q}^{-1}(\cdot)$, it is by definition a bijection function, i.e., given the index $\bm{U}$ or the reconstruction $\hat{\bm{X}}$, we can uniquely determine the corresponding value of $\hat{\bm{X}}$ or $\bm{U}$, respectively. Therefore, we have
\begin{equation}\label{eq:proof_lemma1_eq1}
p(\hat{\bm{X}}|\bm{U})=
\left\{
     \begin{array}{lr}
     1, \quad \text{if}\ \ \hat{\bm{X}}={\rm Q}^{-1}(\bm{U})  \\
     0, \quad \text{if}\ \ \hat{\bm{X}}\neq{\rm Q}^{-1}(\bm{U}) 
     \end{array}
\right.,
\end{equation}
\begin{equation}
p(\bm{U}|\hat{\bm{X}})=
\left\{
     \begin{array}{lr}
     1, \quad \text{if}\ \ \bm{U}={\rm Q}(\hat{\bm{X}})  \\
     0, \quad \text{if}\ \ \bm{U}\neq{\rm Q}(\hat{\bm{X}}) 
     \end{array}
\right.,
\end{equation}
and overall, 
\begin{equation}\label{eq:proof_lemma1_eq3}
p(\hat{\bm{X}}|\bm{X})=
\left\{
     \begin{array}{lr}
     1, \quad \hat{\bm{X}}={\rm Q}^{-1}({\rm Q}(\bm{X}))  \\
     0, \quad \hat{\bm{X}}\neq{\rm Q}^{-1}({\rm Q}(\bm{X}))
     \end{array}
\right..
\end{equation}
Thus, given Eqn.(\ref{eq:proof_lemma1_eq1}) to Eqn.(\ref{eq:proof_lemma1_eq3}), the following conditional entropy is by definition equal to 0:
\begin{equation}\label{eq:proof_lemma1_eq4}
H(\bm{\hat{X}}|\bm{U})=H(\bm{U}|\bm{\hat{X}})=H(\bm{\hat{X}}|\bm{X})=0.
\end{equation}
Furthermore, from the equality of mutual information $I(\bm{X};\bm{\hat{X}})$ and the chain rule of joint entropy $H(\bm{\hat{X}},\bm{U})$, we have
\begin{equation}\label{eq:proof_lemma1_eq5}
\left\{
     \begin{array}{lr}
     I(\bm{X};\bm{\hat{X}}) = H(\bm{\hat{X}}) - H(\bm{\hat{X}}|\bm{X}) \\
     H(\bm{U}|\bm{\hat{X}}) + H(\bm{\hat{X}}) = H(\bm{\hat{X}}|\bm{U}) + H(\bm{U})
     \end{array}
\right..
\end{equation}
Substituting Eqn.(\ref{eq:proof_lemma1_eq4}) into Eqn.(\ref{eq:proof_lemma1_eq5}), we conclude 
\begin{equation}
\left\{
     \begin{array}{lr}
     I(\bm{X};\bm{\hat{X}}) = H(\bm{\hat{X}}) \\
     H(\bm{\hat{X}}) = H(\bm{U})
     \end{array}
\right.,
\end{equation}
and thus
\begin{gather}
     H(\bm{U}) = I(\bm{X};\bm{\hat{X}}).
\end{gather}

\section{Proof of Lemma~\ref{lemma:trans-eqn}}\label{sec:Proof-of-Lemma2}
Recalling the proof in~\cref{sec:Proof-of-Lemma1}, since both the analysis transform $\mathrm{T}_\mathrm{A}(\cdot)$ and synthesis transform $\mathrm{T}_\mathrm{S}(\cdot)$ are deterministic, the following holds:
\begin{equation}
p(\bm{U}|\bm{X})=
\left\{
     \begin{array}{lr}
     1, \quad \text{if}\ \ \bm{U}={\rm Q}(\mathrm{T}_\mathrm{A}(\bm{X}))  \\
     0, \quad \text{if}\ \ \bm{U}\neq{\rm Q}(\mathrm{T}_\mathrm{A}(\bm{X})) 
     \end{array}
\right.,
\end{equation}
\begin{equation}
p(\hat{\bm{X}}|\bm{U})=
\left\{
     \begin{array}{lr}
     1, \quad \text{if}\ \ \hat{\bm{X}}=\mathrm{T}_\mathrm{S}({\rm Q}^{-1}(\bm{U}))  \\
     0, \quad \text{if}\ \ \hat{\bm{X}}\neq\mathrm{T}_\mathrm{S}({\rm Q}^{-1}(\bm{U}))
     \end{array}
\right.,
\end{equation}
\begin{equation}
p(\hat{\bm{X}}|\bm{X})=
\left\{
     \begin{array}{lr}
     1, \quad \text{if}\ \ \hat{\bm{X}}=\mathrm{T}_\mathrm{S}({\rm Q}^{-1}({\rm Q}(\mathrm{T}_\mathrm{A}(\bm{X}))))  \\
     0, \quad \text{if}\ \ \hat{\bm{X}}\neq\mathrm{T}_\mathrm{S}({\rm Q}^{-1}({\rm Q}(\mathrm{T}_\mathrm{A}(\bm{X}))))
     \end{array}
\right.,
\end{equation}
and therefore
\begin{equation}\label{eq:proof_lemma2_eq1}
H(\bm{\hat{X}}|\bm{U})=H(\bm{\hat{X}}|\bm{X})=0.
\end{equation}
Herein, the core distinction between the transform coding and direct coding models lies in the fact that the synthesis transform $\mathrm{T}_\mathrm{S}(\cdot)$, unlike the dequantization function ${\rm Q}^{-1}(\cdot)$, does not inherently guarantee a bijective mapping, particularly in the context of neural transforms. Consequently, given the reconstruction $\hat{\bm{X}}$, there may be uncertainty in $\bm{\hat{Y}}$ and thus $\bm{U}$, i.e.,
\begin{equation}\label{eq:proof_lemma2_eq2}
H(\bm{U}|\bm{\hat{X}})\neq0.
\end{equation}
Substituting Eqn.(\ref{eq:proof_lemma2_eq1}) and Eqn.(\ref{eq:proof_lemma2_eq2}) into Eqn.(\ref{eq:proof_lemma1_eq5}), we can conclude 
\begin{equation}
\left\{
     \begin{array}{lr}
     I(\bm{X};\bm{\hat{X}}) = H(\bm{\hat{X}}) \\
     H(\bm{\hat{X}}) = H(\bm{U}) - H(\bm{U}|\bm{\hat{X}})
     \end{array}
\right.,
\end{equation}
and thus
\begin{gather}
     H(\bm{U}) = I(\bm{X};\bm{\hat{X}})+H(\bm{U}|\bm{\hat{X}}).
\end{gather}

\begin{figure*}[h]
\centering
\includegraphics[width=0.98\textwidth]{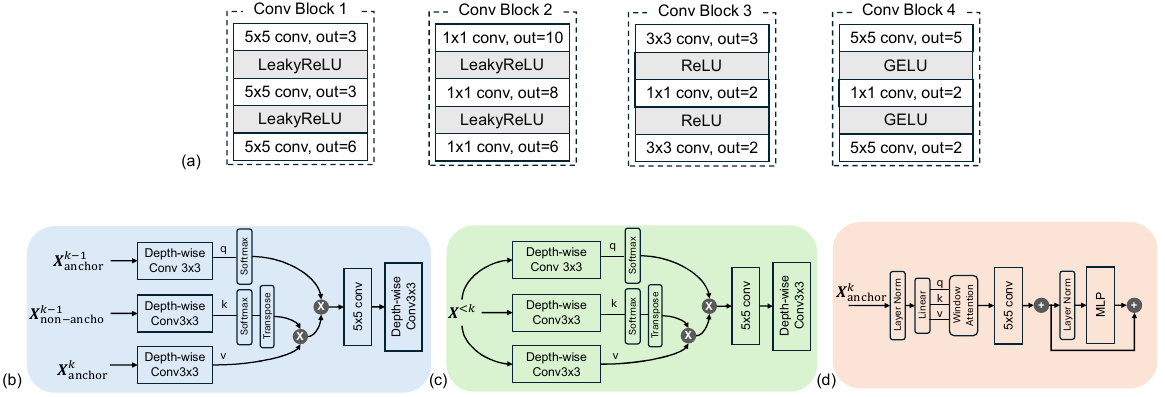}
\caption{Details of (a) Four convolutional modules. (b) Inter attention module, (c) Intra attention module, and (d) Checkerboard attention module from MLIC++~\cite{jiang2023mlic}.}
\label{fig.cond-source-model-blocks}
\end{figure*}

\begin{figure}[h]
\centering
\includegraphics[width=0.40\textwidth]{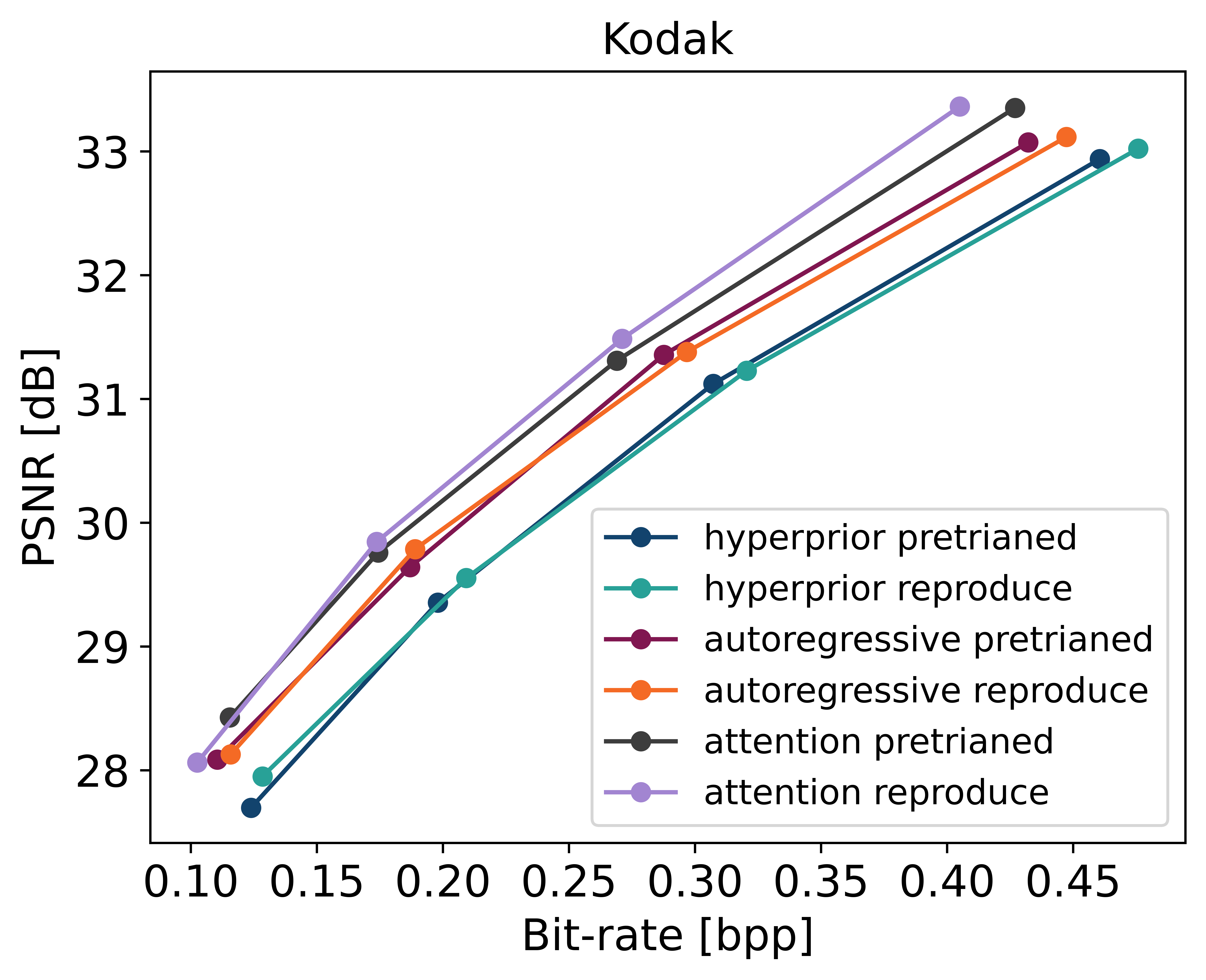}
\caption{Comparison between the pre-trained models from CompressAI’s and our reproduction on the Kodak dataset. Our reproduction achieves 0.52\%, 0.56\%, and -3.32\% BD-Rates for the \textit{hyperprior}, \textit{autoregressive}, and \textit{attention} models, respectively.}
\label{fig.200Wstep-pretrain-Kodak}
\end{figure}

\section{Reproduced baselines}\label{sec:Reproduction}
We retrain the \textit{hyperprior}~\cite{balle2018variational}, \textit{autoregressive}~\cite{minnen2018joint} and \textit{attention}~\cite{cheng2020learned} models from scratch, adhering to the default implementation and training configurations of CompressAI's~\cite{begaint2020compressai}. Four bit-rate points, i.e., $\lambda\in\{0.0018, 0.0035, 0.0067, 0.0130\}$ are trained with $2\times10^{6}$ steps. The evaluation results are summarized in Table~\ref{tab:Reproduced}. On average, when compared to the pre-trained models of CompressAI's, our reproduction models yield 0.39\%, 0.41\%, and -4.11\% BD-Rates~\cite{bjontegaard2001calculation} for the \textit{hyperprior}, \textit{autoregressive}, and \textit{attention} models, respectively. Notably, on the CLIC dataset, our reproduction models outperform their pre-trained counterparts by -1.24\%, -1.56\%, and -5.68\% BD-Rates for the \textit{hyperprior}, \textit{autoregressive}, and \textit{attention} models, respectively. For the \textit{attention} model, our reproduction model outperforms the pre-trained model on all three evaluation datasets of Kodak, CLIC, and Tecnick by -3.32\%, -3.33\%, and -5.68\% BD-Rates, respectively. Note that in the original \textit{attention} network~\cite{cheng2020learned}, the latent is modeled as the mixture of Gaussian with 3 clusters. In CompressAI's default implementation, this cluster number is simplified to 1. We follow this latent implementation. The detailed rate-distortion curves for the Kodak dataset are visualized in Fig.\ref{fig.200Wstep-pretrain-Kodak}.

\begin{table}[htbp]
  \centering
  \caption{BD-Rate comparison between our reproduction and the pre-trained models.}
    \begin{tabular}{ccccc}
    \toprule
          & Kodak & Tecnick & CLIC  & Average \\
    \midrule
    \textit{hyperprior} & 0.52\%  & 1.88\%  & -1.24\% & 0.39\% \\
    \midrule
    \textit{autoregressive} & 0.56\%  & 2.25\%  & -1.56\% & 0.41\% \\
    \midrule
    \textit{attention} & -3.32\% & -3.33\% & -5.68\% & -4.11\% \\
    \bottomrule
    \end{tabular}%
  \label{tab:Reproduced}%
\end{table}%

\section{More details on source entropy models}\label{sec:source-model}
Herein, the module designs are identical to the latent designs~\cite{balle2018variational,minnen2018joint,cheng2020learned,he2022elic,jiang2023mlic}, with only minor dimension adjustment. The details are depicted in Fig.\ref{fig.cond-source-model-blocks}. For the attention modules, the depthwise separable convolution is adopted~\cite{chollet2017xception}.

%% file: main.bbl
\begin{thebibliography}{10}

\bibitem{shannon1959coding}
C.~E. Shannon {\em et~al.}, ``Coding theorems for a discrete source with a fidelity criterion,'' {\em IRE Nat. Conv. Rec}, vol.~4, no.~142-163, p.~1, 1959.

\bibitem{sayood2017introduction}
K.~Sayood, {\em Introduction to data compression}.
\newblock Morgan Kaufmann, 2017.

\bibitem{yang2023introduction}
Y.~Yang, S.~Mandt, L.~Theis, {\em et~al.}, ``An introduction to neural data compression,'' {\em Foundations and Trends{\textregistered} in Computer Graphics and Vision}, vol.~15, no.~2, pp.~113--200, 2023.

\bibitem{balle2018variational}
J.~Ball{\'e}, D.~Minnen, S.~Singh, S.~J. Hwang, and N.~Johnston, ``Variational image compression with a scale hyperprior,'' {\em arXiv preprint arXiv:1802.01436}, 2018.

\bibitem{berger2003rate}
T.~Berger, ``Rate-distortion theory,'' {\em Wiley Encyclopedia of Telecommunications}, 2003.

\bibitem{goyal2001theoretical}
V.~K. Goyal, ``Theoretical foundations of transform coding,'' {\em IEEE Signal Processing Magazine}, vol.~18, no.~5, pp.~9--21, 2001.

\bibitem{balle2016end}
J.~Ball{\'e}, V.~Laparra, and E.~P. Simoncelli, ``End-to-end optimized image compression,'' {\em arXiv preprint arXiv:1611.01704}, 2016.

\bibitem{minnen2018joint}
D.~Minnen, J.~Ball{\'e}, and G.~D. Toderici, ``Joint autoregressive and hierarchical priors for learned image compression,'' {\em Advances in neural information processing systems}, vol.~31, 2018.

\bibitem{minnen2020channel}
D.~Minnen and S.~Singh, ``Channel-wise autoregressive entropy models for learned image compression,'' in {\em 2020 IEEE International Conference on Image Processing (ICIP)}, pp.~3339--3343, IEEE, 2020.

\bibitem{cheng2020learned}
Z.~Cheng, H.~Sun, M.~Takeuchi, and J.~Katto, ``Learned image compression with discretized gaussian mixture likelihoods and attention modules,'' in {\em Proceedings of the IEEE/CVF conference on computer vision and pattern recognition}, pp.~7939--7948, 2020.

\bibitem{he2021checkerboard}
D.~He, Y.~Zheng, B.~Sun, Y.~Wang, and H.~Qin, ``Checkerboard context model for efficient learned image compression,'' in {\em Proceedings of the IEEE/CVF Conference on Computer Vision and Pattern Recognition}, pp.~14771--14780, 2021.

\bibitem{jiang2023mlic}
W.~Jiang and R.~Wang, ``Mlic++: Linear complexity multi-reference entropy modeling for learned image compression,'' in {\em ICML 2023 Workshop Neural Compression: From Information Theory to Applications}, 2023.

\bibitem{qin2024mambavc}
S.~Qin, J.~Wang, Y.~Zhou, B.~Chen, T.~Luo, B.~An, T.~Dai, S.~Xia, and Y.~Wang, ``Mambavc: Learned visual compression with selective state spaces,'' {\em arXiv preprint arXiv:2405.15413}, 2024.

\bibitem{zhang2023uniform}
H.~Zhang, L.~Li, and D.~Liu, ``On uniform scalar quantization for learned image compression,'' {\em arXiv preprint arXiv:2309.17051}, 2023.

\bibitem{agustsson2017soft}
E.~Agustsson, F.~Mentzer, M.~Tschannen, L.~Cavigelli, R.~Timofte, L.~Benini, and L.~V. Gool, ``Soft-to-hard vector quantization for end-to-end learning compressible representations,'' {\em Advances in neural information processing systems}, vol.~30, 2017.

\bibitem{yang2020improving}
Y.~Yang, R.~Bamler, and S.~Mandt, ``Improving inference for neural image compression,'' {\em Advances in Neural Information Processing Systems}, vol.~33, pp.~573--584, 2020.

\bibitem{agustsson2020universally}
E.~Agustsson and L.~Theis, ``Universally quantized neural compression,'' {\em Advances in neural information processing systems}, vol.~33, pp.~12367--12376, 2020.

\bibitem{guo2021soft}
Z.~Guo, Z.~Zhang, R.~Feng, and Z.~Chen, ``Soft then hard: Rethinking the quantization in neural image compression,'' in {\em International Conference on Machine Learning}, pp.~3920--3929, PMLR, 2021.

\bibitem{kingma2019introduction}
D.~P. Kingma, M.~Welling, {\em et~al.}, ``An introduction to variational autoencoders,'' {\em Foundations and Trends{\textregistered} in Machine Learning}, vol.~12, no.~4, pp.~307--392, 2019.

\bibitem{zhao2018adversarially}
J.~Zhao, Y.~Kim, K.~Zhang, A.~Rush, and Y.~LeCun, ``Adversarially regularized autoencoders,'' in {\em International conference on machine learning}, pp.~5902--5911, PMLR, 2018.

\bibitem{xu2020learning}
H.~Xu, D.~Luo, R.~Henao, S.~Shah, and L.~Carin, ``Learning autoencoders with relational regularization,'' in {\em International Conference on Machine Learning}, pp.~10576--10586, PMLR, 2020.

\bibitem{wu2020vector}
H.~Wu and M.~Flierl, ``Vector quantization-based regularization for autoencoders,'' in {\em Proceedings of the AAAI Conference on Artificial Intelligence}, vol.~34, pp.~6380--6387, 2020.

\bibitem{le2018supervised}
L.~Le, A.~Patterson, and M.~White, ``Supervised autoencoders: Improving generalization performance with unsupervised regularizers,'' {\em Advances in neural information processing systems}, vol.~31, 2018.

\bibitem{ma2018constrained}
T.~Ma, J.~Chen, and C.~Xiao, ``Constrained generation of semantically valid graphs via regularizing variational autoencoders,'' {\em Advances in Neural Information Processing Systems}, vol.~31, 2018.

\bibitem{sinha2021consistency}
S.~Sinha and A.~B. Dieng, ``Consistency regularization for variational auto-encoders,'' {\em Advances in Neural Information Processing Systems}, vol.~34, pp.~12943--12954, 2021.

\bibitem{ochoa2019discrete}
H.~Ochoa-Dominguez and K.~R. Rao, {\em Discrete cosine transform}.
\newblock CRC Press, 2019.

\bibitem{toderici2015variable}
G.~Toderici, S.~M. O'Malley, S.~J. Hwang, D.~Vincent, D.~Minnen, S.~Baluja, M.~Covell, and R.~Sukthankar, ``Variable rate image compression with recurrent neural networks,'' {\em arXiv preprint arXiv:1511.06085}, 2015.

\bibitem{liu2019non}
H.~Liu, T.~Chen, P.~Guo, Q.~Shen, X.~Cao, Y.~Wang, and Z.~Ma, ``Non-local attention optimized deep image compression,'' {\em arXiv preprint arXiv:1904.09757}, 2019.

\bibitem{lu2021transformer}
M.~Lu, P.~Guo, H.~Shi, C.~Cao, and Z.~Ma, ``Transformer-based image compression,'' {\em arXiv preprint arXiv:2111.06707}, 2021.

\bibitem{liu2023learned}
J.~Liu, H.~Sun, and J.~Katto, ``Learned image compression with mixed transformer-cnn architectures,'' in {\em Proceedings of the IEEE/CVF conference on computer vision and pattern recognition}, pp.~14388--14397, 2023.

\bibitem{ma2020end}
H.~Ma, D.~Liu, N.~Yan, H.~Li, and F.~Wu, ``End-to-end optimized versatile image compression with wavelet-like transform,'' {\em IEEE Transactions on Pattern Analysis and Machine Intelligence}, vol.~44, no.~3, pp.~1247--1263, 2020.

\bibitem{zhang2023lvqac}
X.~Zhang and X.~Wu, ``Lvqac: Lattice vector quantization coupled with spatially adaptive companding for efficient learned image compression,'' in {\em Proceedings of the IEEE/CVF Conference on Computer Vision and Pattern Recognition}, pp.~10239--10248, 2023.

\bibitem{feng2023nvtc}
R.~Feng, Z.~Guo, W.~Li, and Z.~Chen, ``Nvtc: Nonlinear vector transform coding,'' in {\em Proceedings of the IEEE/CVF Conference on Computer Vision and Pattern Recognition}, pp.~6101--6110, 2023.

\bibitem{suhring2022trellis}
K.~S{\"u}hring, M.~Sch{\"a}fer, J.~Pfaff, H.~Schwarz, D.~Marpe, and T.~Wiegand, ``Trellis-coded quantization for end-to-end learned image compression,'' in {\em 2022 IEEE International Conference on Image Processing (ICIP)}, pp.~3306--3310, IEEE, 2022.

\bibitem{ge2024nlic}
Z.~Ge, S.~Ma, W.~Gao, J.~Pan, and C.~Jia, ``Nlic: Non-uniform quantization based learned image compression,'' {\em IEEE Transactions on Circuits and Systems for Video Technology}, 2024.

\bibitem{guo2021causal}
Z.~Guo, Z.~Zhang, R.~Feng, and Z.~Chen, ``Causal contextual prediction for learned image compression,'' {\em IEEE Transactions on Circuits and Systems for Video Technology}, vol.~32, no.~4, pp.~2329--2341, 2021.

\bibitem{zhu2022unified}
X.~Zhu, J.~Song, L.~Gao, F.~Zheng, and H.~T. Shen, ``Unified multivariate gaussian mixture for efficient neural image compression,'' in {\em Proceedings of the IEEE/CVF Conference on Computer Vision and Pattern Recognition}, pp.~17612--17621, 2022.

\bibitem{fu2023learned}
H.~Fu, F.~Liang, J.~Lin, B.~Li, M.~Akbari, J.~Liang, G.~Zhang, D.~Liu, C.~Tu, and J.~Han, ``Learned image compression with gaussian-laplacian-logistic mixture model and concatenated residual modules,'' {\em IEEE Transactions on Image Processing}, vol.~32, pp.~2063--2076, 2023.

\bibitem{hu2020coarse}
Y.~Hu, W.~Yang, and J.~Liu, ``Coarse-to-fine hyper-prior modeling for learned image compression,'' in {\em Proceedings of the AAAI Conference on Artificial Intelligence}, vol.~34, pp.~11013--11020, 2020.

\bibitem{theis2017lossy}
L.~Theis, W.~Shi, A.~Cunningham, and F.~Husz{\'a}r, ``Lossy image compression with compressive autoencoders,'' {\em arXiv preprint arXiv:1703.00395}, 2017.

\bibitem{goodfellow2016deep}
I.~Goodfellow, ``Deep learning,'' 2016.

\bibitem{goodfellow2014generative}
I.~Goodfellow, J.~Pouget-Abadie, M.~Mirza, B.~Xu, D.~Warde-Farley, S.~Ozair, A.~Courville, and Y.~Bengio, ``Generative adversarial nets,'' {\em Advances in neural information processing systems}, vol.~27, 2014.

\bibitem{he2022elic}
D.~He, Z.~Yang, W.~Peng, R.~Ma, H.~Qin, and Y.~Wang, ``Elic: Efficient learned image compression with unevenly grouped space-channel contextual adaptive coding,'' in {\em Proceedings of the IEEE/CVF Conference on Computer Vision and Pattern Recognition}, pp.~5718--5727, 2022.

\bibitem{begaint2020compressai}
J.~B{\'e}gaint, F.~Racap{\'e}, S.~Feltman, and A.~Pushparaja, ``Compressai: a pytorch library and evaluation platform for end-to-end compression research,'' {\em arXiv preprint arXiv:2011.03029}, 2020.

\bibitem{jiang2022unofficialelic}
W.~Jiang, ``Unofficial elic.'' \url{https://github.com/JiangWeibeta/ELIC}, 2022.

\bibitem{liu2020unified}
J.~Liu, G.~Lu, Z.~Hu, and D.~Xu, ``A unified end-to-end framework for efficient deep image compression,'' {\em arXiv preprint arXiv:2002.03370}, 2020.

\bibitem{kingma2014adam}
D.~P. Kingma, ``Adam: A method for stochastic optimization,'' {\em arXiv preprint arXiv:1412.6980}, 2014.

\bibitem{kodak1993kodak}
E.~Kodak, ``Kodak lossless true color image suite (photocd pcd0992),'' {\em URL http://r0k.us/graphics/kodak}, vol.~6, p.~2, 1993.

\bibitem{CLIC}
``{6th Challenge on Learned Image Compression}.'' \url{https://www.compression.cc/tasks/index.html}.
\newblock Accessed: 2024-11-13.

\bibitem{asuni2014testimages}
N.~Asuni and A.~Giachetti, ``Testimages: a large-scale archive for testing visual devices and basic image processing algorithms.,'' in {\em STAG}, pp.~63--70, 2014.

\bibitem{zhang2024few}
T.~Zhang, H.~Zhang, Y.~Li, L.~Li, and D.~Liu, ``Few-shot domain adaptation for learned image compression,'' {\em arXiv preprint arXiv:2409.11111}, 2024.

\bibitem{lv2023dynamic}
Y.~Lv, J.~Xiang, J.~Zhang, W.~Yang, X.~Han, and W.~Yang, ``Dynamic low-rank instance adaptation for universal neural image compression,'' in {\em Proceedings of the 31st ACM International Conference on Multimedia}, pp.~632--642, 2023.

\bibitem{yang2021implicit}
J.~Yang, S.~Shen, H.~Yue, and K.~Li, ``Implicit transformer network for screen content image continuous super-resolution,'' {\em Advances in Neural Information Processing Systems}, vol.~34, pp.~13304--13315, 2021.

\bibitem{min2017unified}
X.~Min, K.~Ma, K.~Gu, G.~Zhai, Z.~Wang, and W.~Lin, ``Unified blind quality assessment of compressed natural, graphic, and screen content images,'' {\em IEEE Transactions on Image Processing}, vol.~26, no.~11, pp.~5462--5474, 2017.

\bibitem{brancati2022bracs}
N.~Brancati, A.~M. Anniciello, P.~Pati, D.~Riccio, G.~Scognamiglio, G.~Jaume, G.~De~Pietro, M.~Di~Bonito, A.~Foncubierta, G.~Botti, {\em et~al.}, ``Bracs: A dataset for breast carcinoma subtyping in h\&e histology images,'' {\em Database}, vol.~2022, p.~baac093, 2022.

\bibitem{chollet2017xception}
F.~Chollet, ``Xception: Deep learning with depthwise separable convolutions,'' in {\em Proceedings of the IEEE conference on computer vision and pattern recognition}, pp.~1251--1258, 2017.

\bibitem{bjontegaard2001calculation}
G.~Bjontegaard, ``Calculation of average {PSNR} differences between {RD}-curves,'' {\em ITU SG16 Doc. VCEG-M33}, 2001.

\end{thebibliography}
